# Non-Contextual Modeling of Sarcasm using a Neural Network Benchmark

**Vinay Ashokkumar** and **N. Dianna Radpour**
Department of Computer Science, Department of Linguistics
State University of New York at Buffalo
{vinayash, diannara}@buffalo.edu

**Abstract**

One of the most crucial components of natural human-robot interaction is artificial intuition and its influence on dialog systems. The intuitive capability that humans have is undeniably extraordinary, and so remains one of the greatest challenges for natural communicative dialogue between humans and robots. In this paper, we introduce a novel probabilistic modeling framework of identifying, classifying and learning features of sarcastic text via training a neural network with human-informed sarcastic benchmarks. This is necessary for establishing a comprehensive sentiment analysis schema that is sensitive to the nuances of sarcasm-ridden text by being trained on linguistic cues. We show that our model provides a good fit for this type of real-world informed data, with potential to achieve as accurate, if not more, than alternatives. Though the implementation and benchmarking is an extensive task, it can be extended via the same method that we present to capture different forms of nuances in communication and making for much more natural and engaging dialogue systems.

## Introduction

### Motivation

Sarcasm has become an increasingly observed nuance in our everyday communication. It primarily exists in the form of ironic or satirical (Riloff et al. 2013) discussion. The use of sarcasm can be seen as having evolved and popularized since the era of online and virtual communication, with its use becoming more common and frequent in conversational settings. This can be evidenced through the study conducted by (Phillips et al. 2015), in which they demonstrated the prevalence of sarcasm in conversation among individuals today, that the current generation far exceeded in ability to identify sarcasm than the older generation. Hence, the area of sarcasm detection within the domains of sentiment analysis, human-computer interaction and opinion mining, is a complicated problem in natural language processing, where creating the best techniques to identify people's opinions expressed in written language is a great challenge with huge potential, as stated by (Farhadloo and Rolland 2016) and (Joshi, Bhattacharyya, and Carman 2016).

Feature-based classification of sarcasm has become one of the prevalent approaches in the realm of natural language processing research , where a variety of lexical, semantic and punctual based features have been tested through a variety of techniques in identifying and classifying sarcastic intent in text. Sarcastic dialogue has been known to be expressed through an exhibition of several written linguistic properties. It is these through properties (which can also be assumed to be features), that the context of irony in which sarcasm is expressed, can be discovered. These features have been analyzed in various studies that have mined data in repositories such as Twitter (Gonzalez-Ibanez, Muresan, and Wacholder 2011), (Filatova 2012) and Amazon (Davidov, Tsur, and Rappoport 2010b).

A point should be noted that in the case of incomplete data or scenarios, the analysis of context through which sarcastic communication can be identified is much harder as with the lack of data and understanding of situational meaning. Such cases include incomplete datasets, noisy or tampered data, or mixed language and partially ciphered text. Therefore, it is useful to gain insight in the textual features present in text depicting sarcasm, so as to improve prediction capability of written sarcastic bodies through the absence of previous context data or in the presence of noise. Consider the two sarcastic sentences below:

*Haha! I'm trying to imagine you with a personality!!*
*God! Aren't we clever??*

In the above two sentences, a careful scrutiny of representation yields the result of identifying and categorizing features that can be linked to reference to sarcasm. This includes the use of certain phrases and words, such as Haha and God, more usage of punctuation (?? and !!), and a difference in reference of person. A classification scheme is be defined to group common instances of features that can be associated with sarcastic text. This scheme is developed by a combination of intuition of understanding of expression of sarcasm from the authors as well as reference to literature of similar work.

### Our Approach

This paper introduces a methodology of using a large amount of Yelp reviews as input feed for the purpose of classifying sarcastic text purely based on non-context based inferences and through only impromptu presence of the text and the features that it holds itself. A variety of parameters that span the semantic and lexical properties of English sentence formation are identified and scrutinized in these reviews. These parameters are chosen based on the success of previous linguistic research that aim to identify the symptoms of sarcasm, as well as the author's' own intuitive understanding of sarcastic text. The features will be explained in a later section of the paper. The method employs minute concepts of parts-of-speech (POS) tagging to recognize certain words that fall under the categorizations of the defined parameters, which is, cumulatively compiled and aggregated as values to be fed into a two layer feed forward multi perceptron network to correctly classify the text as sarcastic or not. Labels that define sarcastic nature of each review are provided by human participation of a group of four people (including the authors), that classify the testing set of reviews. The tagging is employed using an open source python toolkit to natural language processing called NLTK, and the construction of the neural network is also constructed with another open source Python toolkit called TensorFlow. The methodology is repeated for five star segregated sets of reviews, in equal proportions. The reasoning behind this is to validate the authors hypothesis of generalizing sarcasm to be associated with lower rated reviews, which makes it easier to come to a unanimous consensus of marking sarcastic reviews, hence validating our methodology much more accurately. To the best of the authors knowledge, this work is a unique contribution to the field of sentiment analysis, specifically sarcasm modelling that utilizes semantic feature categorization, rather than relying heavily on sentence structure and tagging as in other works.

### Sarcastic Feature Selection

The original focus of the sarcasm problem was in classification. And the research still holds strong to this day. As proposed by (Zhang, Zhang, and Fu 2016), there are two main types of features that are discernible through nature. The first is a scheme which describes a binary based classification to the problem. This scheme is a discrete feature scheme, where individual units of text classify sarcastic text solely on the content and text data that is available to mine. Such features included hashtags and smileys (Davidov, Tsur, and Rappoport 2010a), lexical features (Gonzalez-Ibanez, Muresan, and Wacholder 2011) and language-independent features (Ptacek, Habernal, and Hong 2014). The second scheme analyzes features related to context and emphasizes on phrase context and sentence structure with latent dependencies on grammar rules (Rajadesingan, Zafarani, and Liu 2015); (Bamman and Smith 2015). Work in this area is relatively new and shows promise in terms of classification results. (Rajadesingan, Zafarani, and Liu 2015) also took a new approach in using historical behavioral data to predict the potential of sarcasm in a user's text. Other approaches in feature selection mainly involve neural network methods (Socher et al. 2013), (Dos Santos and Gatti 2014) and (Vo and Zhang 2015).

However, these works mainly focus on scrutinizing different features on their effect on sarcasm analysis through an array of different classification techniques. Furthermore all of the data was centered on Twitter with no novelty in different data sets, which may yield different variables to analyze.

### Review-based Approaches

Work involving analyzing review corpora brings about a different tone of sarcasm that focuses on one subject (product, business, etc.), that allows the consideration of new parameters that can get involved lexically or within the properties of the text to have involvement in sarcastic text. Since this study works on review data, we dedicate a section to go over literature related to sarcasm analysis that used sample data primarily from reviews.

Review text classification has progressed from sentiment analysis research in areas not only in sarcasm, but in detection of irony (Reyes, Rosso, and Veale 2013), satire (Burfoot and Baldwin 2009), and humor (Reyes, Rosso, and Buscaldi 2012). One of the most influential studies in sarcasm detection involved work done by (Tsur, Davidov, and Rappoport 2010), who developed a sophisticated algorithm for sarcasm detection through analyzing a large data set of Amazon product reviews. The study focused on using unique pattern extractions for classification tasks using trained sarcastic labeled corpus for comparison. The approach was through a use of semi-supervised algorithm with a reported F1 score measure of 82%.

(Buschmeier, Cimiano, and Klinger 2014), worked on a review data set of Amazon corpus, earlier published by (Filatova 2012), and extended with user profile reviews from Twitter as well. The research was centered on automatic sarcasm detection, and focused on sentiment irony imbalance between product rating and surface polarity, constructing and classifying features based on this approach. The F1 score measure reported was 74%. This work established baseline comparisons on an Amazon review corpus based on sarcastic notion, generated by (Filatova 2012).

Work done using reviews from Yelp, which is most common to the research proposed in this paper are the studies done by (Bakhshi, Kanuparthy, and Shamma 2014). The focus there was to understand social sentiment as a whole from the reviews. Their findings defined the correlations between the ratings given by users and the tone of positivity or negativity associated with the review. Though the investigation spanned all social signals, it did not concisely focus on the sarcasm environment, thereby ignoring a lot of the prospective inferences that identify ideal conditions for sarcasm.

### Machine Learning Approaches

Sarcasm detection is a relatively new research topic which has gained increasing interest only recently, partly thanks to the rise of social media analytics and sentiment analysis. An early work in this field was done by (Tsur, Davidov, and Rappoport 2010), on a dataset of 6,600 manually annotated Amazon reviews using a kNN-classifier over punctuation- based and pattern-based features, i.e. ordered sequence of high frequency words. (Gonzalez-Ibanez, Muresan, and Wacholder 2011), used support vector machine (SVM) and logistic regression over a feature set of unigrams, dictionary-based lexical features and pragmatic features (e.g., emoticons) and compared the performance of the classifier with that of humans. (Reyes, Rosso, and Veale 2013) described a set of textual features for recognizing irony at a linguistic level, especially in short texts created via Twitter, and constructed a new model that was assessed along two dimensions: representativeness and relevance. (Buschmeier, Cimiano, and Klinger 2014) compared the performance of different classifiers on the Amazon review dataset using the imbalance between the sentiment expressed by the review and the user-given star rating. Features based on frequency (gap between rare and common words), written spoken gap (in terms of difference between usage), synonyms (based on the difference in frequency of synonyms) and ambiguity (number of words with many synonyms) were used by (Barbieri, Saggion, and Ronzano 2014) for sarcasm detection in tweets.

In terms of Bayesian-based approaches, most of the current literature available employs Naive Bayes Classification (NBC) combined with other classification schemes, or where performance has been compared with other classification schemes. A similar themed study was conducted by (Mertiya and Singh 2016), with NBC combined with adjective analysis on movie reviews and tweets. The results yielded a classification accuracy of 82%. (Mukherjee and Bala 2017) combined NBC with Fuzzy Logic to classify sentiment based on features derived from function words and part of speech n-grams. (Itani et al. 2012) demonstrated a performance of NBC with a comparative approach against Naive Search. The study worked with linguistic and lexical features unique to Arabic text. (Dey et al. 2016) extended the comparative approach using NBC and k-Nearest Neighbors (kNN) as competing algorithms. The study proved the effectiveness of using a Bayesian method with NBC outperforming kNN with an overall accuracy of 80%.

## Data

### Dataset

The data set provided is an open source data set in part with the annual competition held by Yelp, called the Yelp Dataset Challenge. The data set consists of information broken down into 5 subsets, classified as data pertaining to type of business related information, reviews, tip information, a biodata of an array of user profiles and information pertaining to check in times.

Numerically summarized, it comprises of 4.1 million reviews, 947,000 tips, aggregated over 125,000 businesses. The demographics span across 4 countries and 11 cities. The data set spans with an updated base ranging from 2005 to 2017. For the purposes of the study of this project, impetus is given only to the review dataset of the entire collection.

### Feature Categorization

This is the stage where the unique properties innate to sarcastic textual expression are categorized for better illustration of the model. The categories are made based on the similarity and same linguistic functionality of related features. The individual features are intuitively chosen from experience in dealing with various Yelp reviews and identifying the sarcastic ones.

The categories are made as follows:

- Keyword features: These are words that express exaggerated emotion, usually emphasized during sarcastic expression of emotion or tone. The basis of the selection of these features are due to their success in Twitter sarcasm detection studies like (Tungthamthiti, Shirai, and Mohd 2014).
- Punctuation features: These are features pertaining to semantic properties of the review text to indicate the tone in which the sentence is being expressed. They were chosen based on the effectiveness in previous works, such as (Davidov, Tsur, and Rappoport 2010b).

- Superlative features: These are words expressed in their highest form of grammatical degree (usually ending in “est”). The hypothesis being that sentences that use superlatives often, mean to indicate higher levels of exaggerated emotion, usually associated with satirical intent.
- Referentiality features: This is a group to measure sentences that express the type of person referenced to in ex- pressing ironical feelings. In this study, words depicting self-reference in the first person are chosen as the distinction to measure referentiality of a sentence in a given review. That is to say, words depicting self referenced, first person views are counted for presence. Examples are ’I’, ’me’, ’myself’, etc.
- Seasonal features: This group of features classify the season of the review of which the sarcastic review is written by month. Months 11, 12 and 1 are considered Winter months. 2, 3 and 4 are Spring. 5, 6, 7, 8 are Summer months and 9, 10 depict Fall. To be easily fed into the neural network, each of the seasons are discretized to values between 0 and 3 to create one hot vectors of size 4.

Table 1: Feature Types of the Model

| Feature | Types |
|---|---|
| Keywords (s) | ”*Wow*”,”*OMG*”,”*Haha*”,”*Damn*” |
| Punctuation | ”!”,”()”,”?”,”...”,””””,”,”,”.”) |
| Referentiality | ”*I*”,”*me*”,”*myself*”,”*we*”,”*us*” |
| Season | ”*Winter*”,”*Fall*”,”*Summer*”,”*Spring* |
| Superlatives | ”*Worst*”,”*Least*”,”*Wordsendingin − est*” |

*Table 1 lists out the parameters (features), as well as the sub-type values they hold in each categorization.*

## Model Construction

### Preprocessing

Due to the limited participation in the study and the strict time constraints involved in the project, a subset of 1000 reviews from each of the five star rating categories are taken from the overall 1.2 million reviews that comprises of the whole dataset. As mentioned earlier in the introduction, the hypothesis is that sarcasm can detected strictly from features described in section ”Feature Categorization”. Take, for example, the sentence:

> *“Wow, not the brightest crayon in the box, now are we?”*

As observed, this sentence can be classified to be sarcastic, and within that there lies features that fall in the defined categories as follows:

- Keyword: Wow
- Punctuation: ?
- Superlative: brightest
- Referential: we

Each review is accompanied by a date with a mm/dd/yyyy format, and the season takes a value based on the definition described previously.

The methodology follows tagging and calculating the occurrence of these features in 1000 reviews extracted from the dataset for each star label, giving a total of 5000 reviews. Prior to that, sarcasm output labels are given to each of the 5000 reviews. A 1 is given if the review is deemed to be sarcastic and a 0 otherwise. There are five participants to label each set of 1000 reviews, which include the authors. After the label is given, the type of each of the defined features along with the aggregated sum of their occurrences appearing in a respective review is stored in a dictionary, with the sarcastic output label given. Occurrences are identified for each kind of feature using the parts-of-speech tagging (POS) implements for each review using the NLTK toolkit in Python. This tags superlatives, punctuations and referential words.

For example, if the review contained two question marks and the use of ’Wow’ twice, the dictionary would contain the following format:

- ‘?’: 2
- ‘WOW’: 2

It is important to note that the the dictionary is case-insensitive for keywords and all cases are considered. The 5000 dictionaries are pickled into five separate pickle files that are based on the star rating of the review. Each pickle file contains the 1000 dictionaries of the same rating, and is then fed into the neural network as described in the next subsection. Figure 1 represents a complete flowchart of the preprocessing stages from extraction to input into the network.

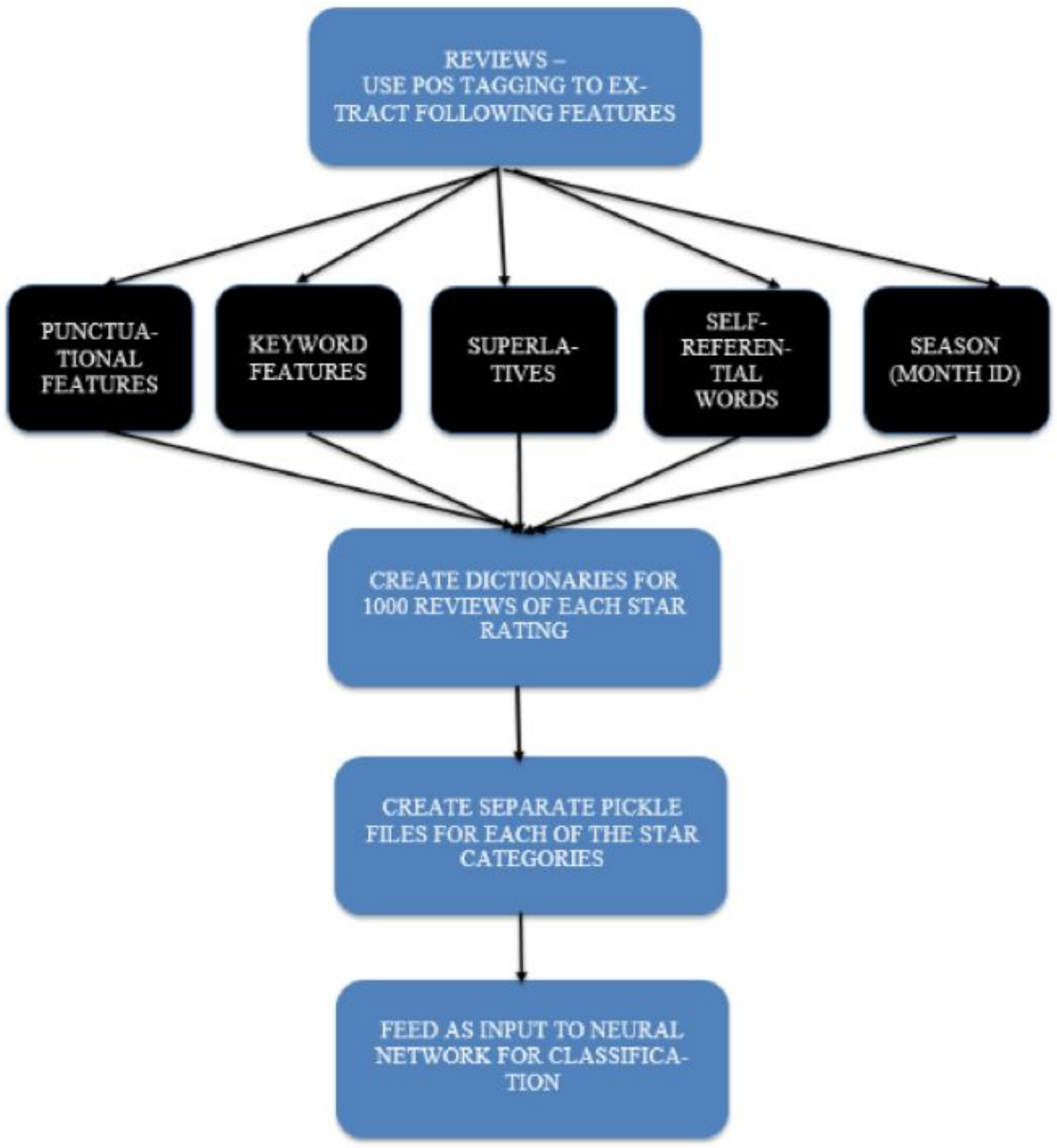

Figure 1: Preprocessing Flowchart

**Network Description**

The training and testing sets for the network comprises of Yelp reviews normalized in the preprocessed form as described in the previous subsection. Out of which, the training set is of size 700, with the testing set of 300 review dictionaries respectively from each of the five pickle files. This is segregated for each of the 5 categories of ratings respectively. As there are 15 main features involved in the dictionary, the number of input nodes formulated for the model is 15. Each of the 700 reviews are fed to the network for each star rating category. From previous study of literature (LeCun and Ranzato 2013), the number of hidden layers of a problem equivalent to the size of the dataset used based on the training and testing dataset sizes is varied between 1 and 2. Each layer consists of 7 to 15 hidden layer nodes. Network parameters such as learning rate and dropout, are varied accordingly. The learning rate is varied from 0.0001 to 0.01 in steps of 0.9 for each testing iteration. The dropout rate remains constant at 0.75.
For optimization, the in-built optimizer provided by TensorFlow, called the Adam Optimizer is employed to calculate the cross entropy between the predicted label and true label for minimizing error. The activation functions used for the input to first hidden layer, and the path from the first hidden layer to the second hidden layer is the activation function called Relu. The normalizing activation function at the output layer chosen is the regular softmax function.

Initially, the network is trained with a dataset of 500 purely sarcastic reviews, handpicked from the author's own judgment in sarcastic understanding. The process is defined as a means of tuning the network to establish thresholds for each of the features for sarcastic benchmarks, in order to gauge the testing set with. The process is repeated with a domination of non-sarcastic reviews in the training set to tune the network in the opposite direction, where a balance is stroked in order for the network to generalize well.

## Experimental Results

Through a tribulation of trial and error, the optimal results for each review to be classified as sarcastic or not yielded the best results with the following network parameters: 1). A learning rate of 0.01. 2). 2 hidden layers. 3). Each hidden layer consisting of 15 hidden nodes. For all 5 star cate- gories of reviews, each of the training set was fed into the network in batches of size 100, cumulatively adding to 7 batches each. At 10 epochs, the optimization was completed for all 5 sets of reviews. The testing set comprised of 300 reviews that were fed into the network unlabeled. An illustration of the highest classification accuracies for all sets of star reviews is given in Figure 2.

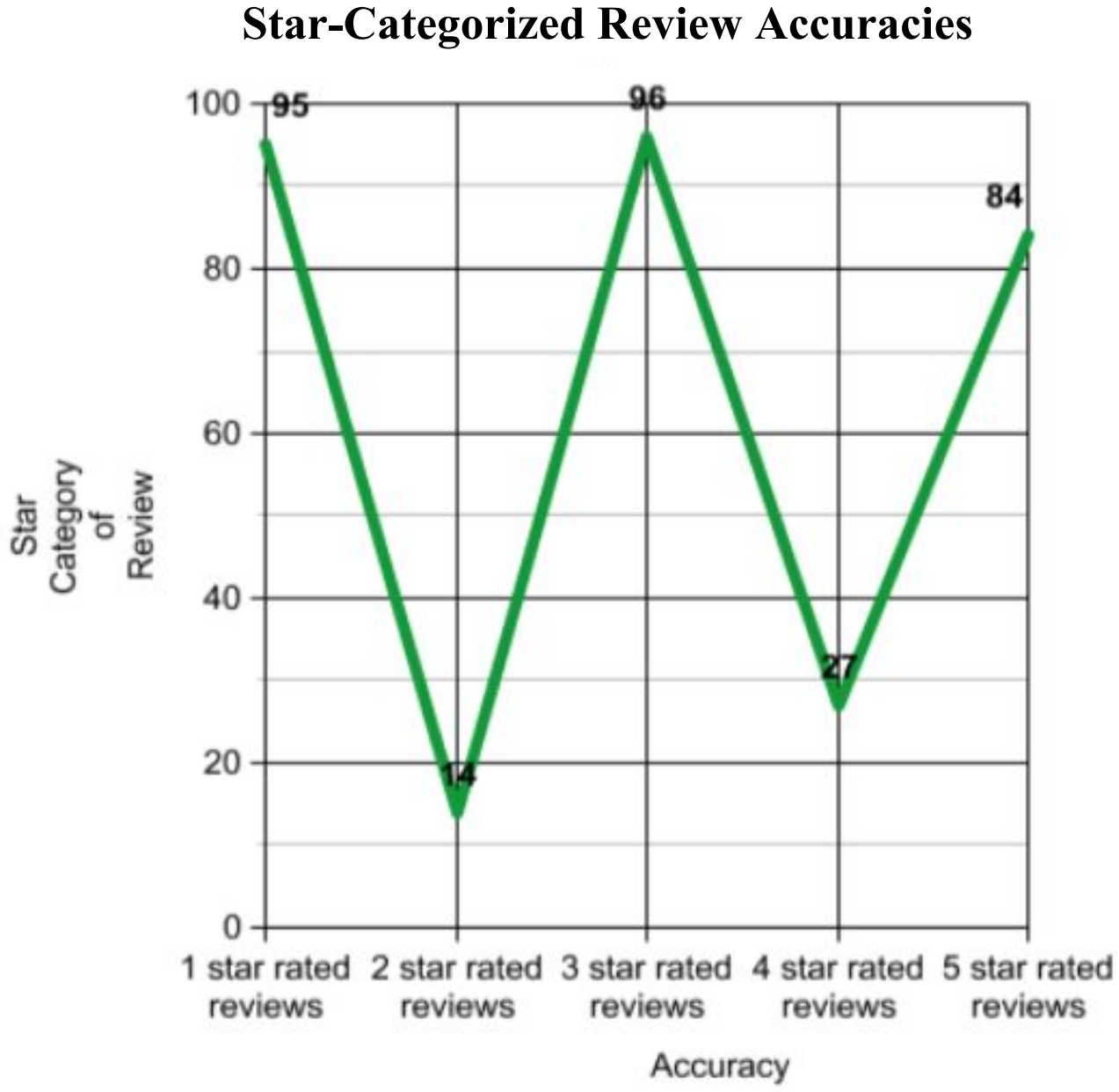

*Figure 2: Star-based Reviews' Sarcasm Accuracies*

As it can be observed, notable results are displayed by 1, 3 and 5-star rated reviews, with the 2 and 4-star reviews receiving generally poor results. A justification for this can be that the high-star rated reviews, there existed more sarcastic reviews in the training set than that of the 2 and 4-star rated reviews. In especially 1 star and 3 star rated reviews, with accuracies of 95% and 96% respectively, there existed an abundance of sarcastic reviews where in 1-star reviews, the tone of sarcasm was in a negative context, and in 5-star rated reviews, the tone of sarcasm was generally in a positive context. Hence, it is plausible that the network was not able to generalize as well in these reviews because the number of sarcastic reviews in these two datasets was extremely minute. 5-star rated reviews demonstrated a nominal performance with 84%, as it evenly balanced out with the network training parameters and hence was able to prevent over fitting with balanced weights that have been learned between the layers of the network. In the dataset used, the 5 star reviews were well balanced with a ratio of slightly more non-sarcastic reviews than sarcastic reviews. The accuracies peaked after 10 epochs, after which they began to fall due to overfitting.

The performance metrics of the classifier are specified in Table 2. The precision and recall for all of the reviews average to 0.68 and 0.71, respectively. The best performance was given by the 2-star reviews, which is a good measure considering most of the there were more than 200 reviews that were deemed sarcastic for both the 1 star and 2-star categories. The final F1 score of the classifier that is averaged from the 5 sets of reviews is 0.68.

Table 2: Performance Metrics for the Classifier Review Based

| Metric | 1 star | 2 star | 3 star | 4 star | 5 star |
|---|---|---|---|---|---|
| Precision | 0.67 | 0.78 | 0.74 | 0.66 | 0.54 |
| Recall | 0.77 | 0.61 | 0.77 | 0.72 | 0.68 |
| F1 | 0.71 | 0.68 | 0.75 | 0.69 | 0.61 |

## Conclusion

Human-robot dialogue could be greatly enhanced using this approach of benchmarking the neural networks with values tabulated from real-world information. It is a simple, yet effective technique, with in numerous implications. Our proposed method aids accurately gauging a given sentence or phrase as sarcastic or not without much priori knowledge of situational context or intent, making it a potential solution to sparse data problems. This paired with audio data and preprocessing data on acoustic cues to match the annotated sentiments of the reviews would have far-reaching implications for natural language user interfaces, particularly in establishing conversational agents with artificial intuition that are sensitive to phonetic nuances when interacting with humans.

## Future Work

There exists some limitations in the methodology of this study, such as improper division of sarcastic and non-sarcastic reviews in each star category, analysis with more lexical based features and use of more than just the authors consensus in understanding to label reviews are sarcastic or not, and involve a much larger size of an opinion pool to provide sarcastic labeling. Such limitations shall be addressed in future work, as well as the implementation of different machine learning models, like Restricted Boltzmann Machines and Hidden Markov Models to generate a time series model to analysis features of sentence so as to identify when a sentence becomes sarcastic or not in real time.